\pdfoutput=1

\documentclass[11pt]{article}

\usepackage[review]{acl}

\usepackage{times}
\usepackage{latexsym}

\usepackage[T1]{fontenc}

\usepackage[utf8]{inputenc}

\usepackage{microtype}

\usepackage{graphicx}
\usepackage{float}
\title{Context-Dependent Embedding Utterance Representations for Emotion Recognition in Conversations}

\author{Patrícia Pereira\textsuperscript{1,2}, Helena Moniz\textsuperscript{1,3}, Isabel Dias\textsuperscript{1,2} \and  Joao Paulo Carvalho\textsuperscript{1,2} \\
\textsuperscript{1}INESC-ID, Lisbon \\
\textsuperscript{2}Instituto Superior Técnico, University of Lisbon \\
\textsuperscript{3}Faculdade de Letras, University of Lisbon \\
\texttt{\{patriciaspereira, isabel.h.dias\}@tecnico.ulisboa.pt} \\
\texttt{\{helena.moniz, joao.carvalho\}@inesc-id.pt} \\
}

\begin{document}
\nolinenumbers
\makeatletter\acl@finalcopytrue
\maketitle
\begin{abstract}
Emotion Recognition in Conversations (ERC) has been gaining increasing importance as conversational agents become more and more common. Recognizing emotions is key for effective communication, being a crucial component in the development of effective and empathetic conversational agents. Knowledge and understanding of the conversational context are extremely valuable for identifying the emotions of the interlocutor. We thus approach Emotion Recognition in Conversations leveraging the conversational context, i.e., taking into attention previous conversational turns. The usual approach to model the conversational context has been to produce context-independent representations of each utterance and subsequently perform contextual modeling of these. Here we propose context-dependent embedding representations of each utterance by leveraging the contextual representational power of pre-trained transformer language models. In our approach, we feed the conversational context appended to the utterance to be classified as input to the RoBERTa encoder, to which we append a simple classification module, thus discarding the need to deal with context after obtaining the embeddings since these constitute already an efficient representation of such context. We also investigate how the number of introduced conversational turns influences our model performance. The effectiveness of our approach is validated on the open-domain DailyDialog dataset and on the task-oriented EmoWOZ dataset.
\end{abstract}

\begin{figure}[!ht]
\begin{center}
  \includegraphics[width=0.9\linewidth]{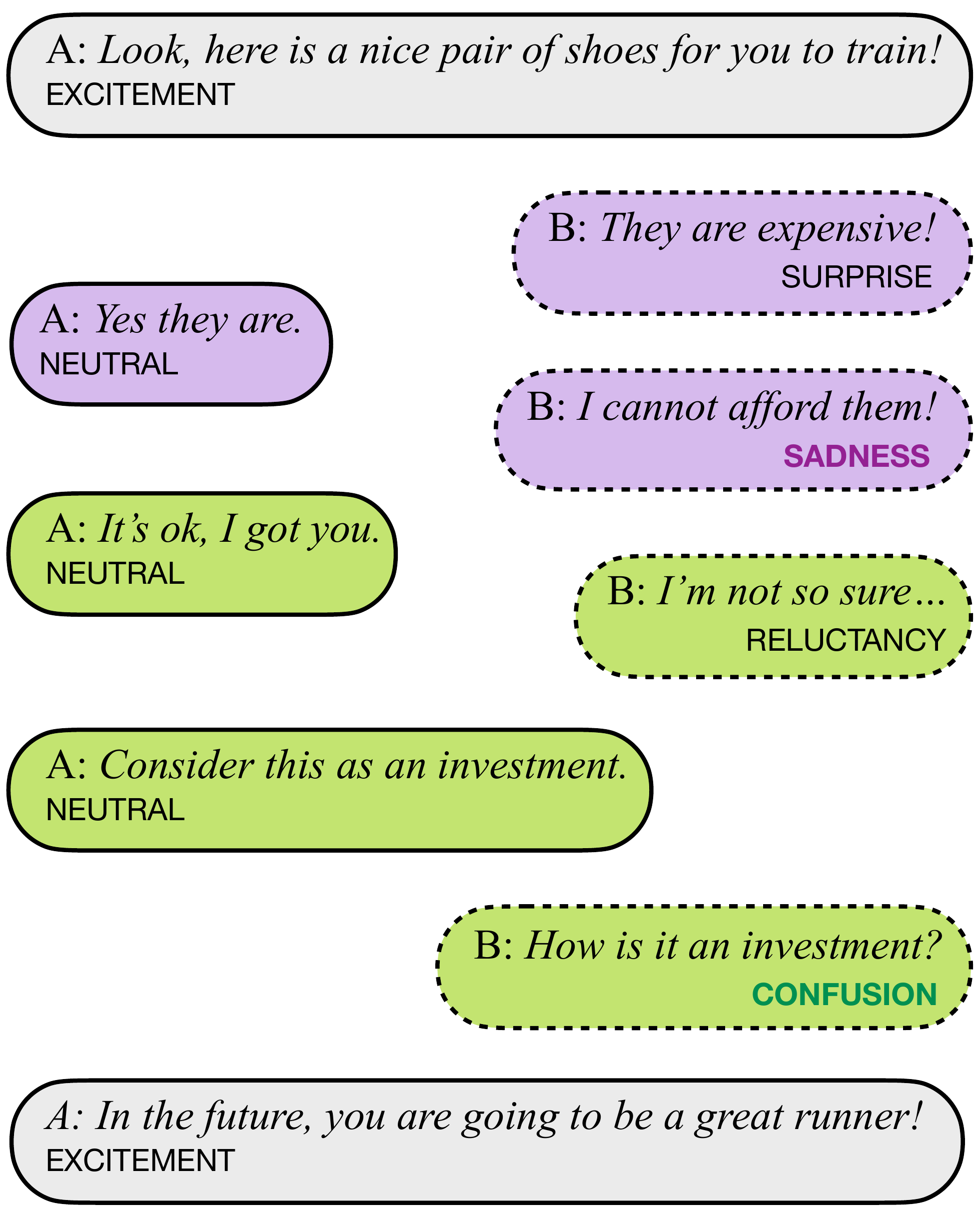}
  
 \end{center}
  \caption{A dialogue in which context is key to infer the associated emotions. To infer the emotions \emph{Sadness} and  \emph{Confusion}, knowledge of the present and previous two and three utterances is required, respectively.
}
  \label{f}
  \end{figure}

\section{Introduction}
Emotion Recognition in Conversations (ERC) is useful in automatic
opinion mining, emotion-aware conversational agents
and assisting modules for therapeutic practices. There is thus an increasing interest in endowing machines with efficient emotion recognition modules.

Knowledge and understanding of the conversational context, i.e., of the previous conversation turns, are extremely valuable in identifying the emotions of the interlocutors \cite{poria2019emotion} \cite{chatterjee2019semeval} \cite{pereira2022deep}.

Research in automatic emotion recognition using machine learning techniques dates back to the end of the 20th century. 
However, the use of the conversational context as an auxiliary information for the classifiers, did not appear until publicly available conversational datasets became more common.

State-of-the-art ERC works leverage not only state-of-the-art pre-trained-language models such as BERT \cite{devlin2018bert} and RoBERTa \cite{liu2019roberta}, but also deep, complex architectures to model several factors that influence the emotions in the conversation \cite{pereira2022deep}. Such factors usually pertain to self and inter-speaker emotional influence and the context and emotion of preceeding utterances.

In this paper we argue that the powerful representation capabilities of pre-trained language models can be leveraged to model context without the need of additional elaborate classifier architectures, allowing for much simpler and efficient architectures. Furthermore, it is our contention that the Transformer, the backbone of our chosen language model, is better at preserving the contextual information since it has a shorter path of information flow than the RNNs typically used for context modelling. In this line, we rely on the RoBERTa language model and resort to a simple classification module to preserve the contextual information. 

The usual approach to model the conversational context has been to produce context independent representations of each utterance and subsequently perform contextual modeling of those representations. State-of-the art approaches start by resorting to embedding representations from language models and employ gated or graph neural network architectures to perform contextual modelling of these embedding representations at a later step. In our much simpler and efficient proposed approach, we produce context-dependent embedding representations of each utterance, by feeding not only the utterance but also its conversational context to the language model. We thus discard the need to deal with context after obtaining the embeddings since these constitute already an efficient representation of such context.

Our experiments show that by leveraging context in this way,
one can obtain state-of-the-art results with RoBERTa and a simple classification module, surpassing more complex state-of-the-art models.

\section{Related Work}
\label{rl}
Amongst the first works considering contextual interdependences among utterances is the one by Poria et al. \cite{poria-etal-2017-context}. It uses LSTMs to extract contextual features from the utterances. These gated recurrent networks make it possible to share information between consecutive utterances while preserving its order.

A more elaborate model also leveraging gated recurrent networks is DialogueRNN \cite{majumder2019dialoguernn}, which uses GRUs to model the speaker, context and emotion of preceding utterances by keeping a party state and a global state that are used to model the final emotion representation.

Gated recurrent networks have a long path of information flow which makes it difficult to capture long term dependencies. These can be better captured with the Transformer which a has shorter path of information flow. Its invention in 2017 \cite{vaswani2017attention} led to a new state-of-the-art in several Natural Language Processing tasks. 

Amongst the first works leveraging the Transformer is the Knowledge-Enriched Transformer (KET) \cite{zhong-etal-2019-knowledge}. It uses its self-attention to model context and response. 
It also makes use of an external knowledge base, a graph of concepts that is retrieved for each word.

Following the invention of Transformers, pre-trained language models brought about another new state-of-the art in 2019. Since their invention, most state-of-the art ERC works resorted to encoder pre-trained language models \cite{shen2021dialogxl} \cite{ghosal-etal-2020-cosmic} \cite{li-etal-2021-past-present}.

COSMIC \cite{ghosal-etal-2020-cosmic} leverages RoBERTa Large as feature extractor. Furthermore, it makes use of the commonsense transformer model COMET \cite{bosselut2019comet} in order to extract commonsense features. Five bi-directional GRUs model a context state, internal state, external state, intent state, and emotion state that influence the final emotion classification.

Psychological \cite{li-etal-2021-past-present} also uses RoBERTa Large for utterance encoding and COMET. For conversation-level encoding it constructs a graph of utterances to model the actions and  intentions of the speaker along with the interactions with other utterances. It uses COMET to introduce commonsense knowledge into the graph edge representations and processes this graph using a graph transformer network.

\section{Methodology}

\begin{figure}[!ht]
\begin{center}
  \includegraphics[width=\linewidth]{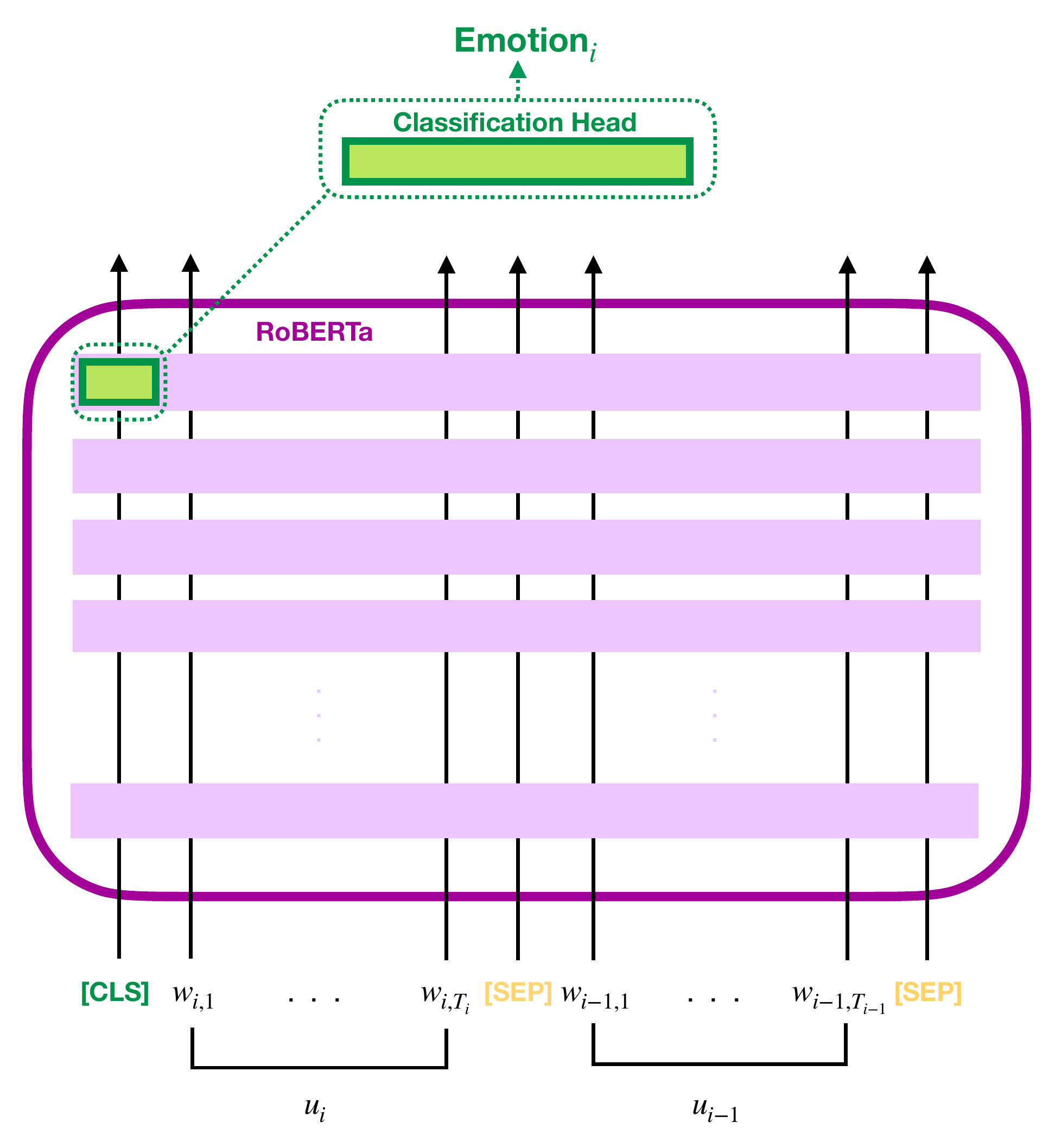}
  
 \end{center}
  \caption{Model architecture. Two utterances are given as input to RoBERTa encoder, of which the CLS token of the last layer is fed to the classification head that predicts the emotion.
}
  \label{f1}
  \end{figure}

We describe how we obtain a contextual embedding representation of the sentence and its context with RoBERTa, how we pool the contextual embeddings, our classification module and how we obtain the emotion labels. These processes can be observed in Figure \ref{f1}.

\subsection{Task definition}

Given a conversation, a sequence of $u_{i}$ utterances with corresponding $emotion_{i}$ from a predefined set of emotions, the aim of the task of ERC is to correctly assign an emotion to each utterance of the conversation.  An utterance consists in a sequence of $w_{it}$ tokens representing its $T_{i}$ words 

\begin{equation}\label{equ}
u_{i}=(w_{i1},w_{i2},...,w_{iT_{i}})
\end{equation}

The usual approach for this task has been to produce context independent representations of each utterance and perform contextual modeling of these. 
In our approach we produce context-dependent representations of each utterance that represent not only the utterance but also a given number of previous utterances from the conversation.

\subsection{Context-dependent feature extraction}

For context-dependent feature extraction, we feed as input to RoBERTa the utterance we intend to classify, $u_{i}$, concatenated with its conversational context corresponding to the number $c$ of previous utterances in the conversation, $(u_{i-1},u_{i-2},...,u_{i-c})$. Concretely, we feed $u_{i}$ to the model, preceded by the [CLS] token and suceded by the [SEP] token, followed by the previous turns $u_{i-1}$ up to $u_{i-c}$, separated by the [SEP] token.



\subsection{Pooling}

The RoBERTa encoder outputs several layers of embeddings representing the utterance, and in our approach, also the preceding utterances it receives as input. Each layer comprises several tokens, being the number of tokens the same as the number of input tokens. Each token is a vector with dimension corresponding to the RoBERTa hidden size.

From these embeddings one can extract a suitable representation for the sentence. Choosing all tokens from all layers would yield an extremely memory demanding classification layer and may not yield the best model performance. Thus we choose the first embedding from the last layer L, the [CLS] which is used for classification, as in Equation \ref{ecls}. 
\begin{equation}\label{ecls}
pooled_{i}=RoBERTa_{L,[CLS]}(input_{i})
\end{equation}

\subsection{Emotion Classification}
The classification module that follows RoBERTa is a linear fully connected layer, applying a linear transformation to the pooled encoder output data. Its input size is the RoBERTa encoder hidden size and its output size is the number of emotion classes.

The final label probability distribution is yielded by applying the softmax operation to the output of the classification head and the predicted label is the one with the highest probability:
\begin{equation}\label{eq6}
emotion_{i}=argmax(Softmax(pooled_{i}W^{T}+b))
\end{equation}

\section{Experimental Setup}

\subsection{Training}

Our model is based on RoBERTa-base from the Transformers library by Hugging Face \cite{wolf-etal-2020-transformers}. It is trained with the cross-entropy loss with logits. The Adam \cite{kingma2014adam} optimizer is used with an initial learning rate of 1e-5 and 5e-5, for the encoder and the classification head, respectively with a layer-wise decay rate of 0.95 after each training epoch for the encoder. The encoder is frozen for the first epoch. The batch size is set to 4. Gradient clipping is set to 1.0. As stopping criteria, early stopping is used to terminate training if there is no improvement after 5 consecutive epochs on the validation set over macro-F1, for a maximum of 10 epochs.  The checkpoint used in testing is the one that achieves the highest macro-F1 score on the validation set.

\subsection{Evaluation}

We evaluate the performance of our model with the macro F1-score. The reported results are yielded from an average of 5 runs corresponding to 5 distinct random seeds that are kept for a meaningful comparison of all experiments. This average is motivated by the fact that results for the same experiment obtained with different random seeds can have a variability of about 3 in macro F1-score which is a large deviation given that our proposed approach yields an improvement of that magnitude and comparison between state-of-the-art models are based on improvements of less than 1 F1-score. This procedure is in line with several authors that also resort to 5 run averages \cite{li-etal-2021-past-present} \cite{zhong-etal-2019-knowledge} \cite{shen2021dialogxl} \cite{shen-etal-2021-directed}.

\vspace{2.5mm}

Our code is publicly available\footnote{\url{http://github.com/patricia-pereira/cd-erc}}.

\subsection{Datasets}

We evaluate our approach on the chit-chat DailyDialog \cite{li2017dailydialog} dataset and on the task-oriented EmoWOZ \cite{feng2022emowoz} dataset.

\subsubsection{DailyDialog}

DailyDialog is built from websites used to practice English dialogue in daily life. It is labelled with the six Ekman’s basic emotions \cite{ekman1999basic}, anger, disgust, fear, happiness, sadness and surprise, or neutral. The publicly available splits of Yanran are used.

\subsubsection{EmoWOZ}

EmoWOZ is derived from MultiWOZ \cite{budzianowski-etal-2018-multiwoz}, one of the largest multi-domain corpora benchmark dataset for various dialogue tasks. User utterances are annotated with either fear, dissatisfaction, apologetic, abusive, excited, satisfied or neutral emotions.

The statistics and proportion of labels in the  datasets are presented in Tables \ref{t0} and \ref{t1}, respectively.

\begin{table}[H]
  \caption{Statistics of the datasets}
 \centering
  \begin{tabular}{cccc}
    \hline
   &  \textbf{DailyDialog} &\textbf{EmoWOZ}  \\
    \hline
    Dlg type & Chit-chat &Task-oriented  \\
    \# Dlgs &13,118&  11,434 \\
    \# Turns &102,979& 167,234 \\
     Avg turns in dlg &7.9&14.6 \\

    \hline
  \end{tabular} 
 \label{t0}
\end{table}

\begin{table}[H]
  \caption{Proportion of labels in the datasets}
 \centering
  \begin{tabular}{cccc}
    \hline
      \multicolumn{4}{c}{\textbf{DailyDialog}}   \\
       \hline
  Ang & Disg & Fear & Hap \\
   
   1.0\%&0.3\%& 0.2\%& 12.5\% \\
    
  & Sad & Sur & Neu \\

  &1.1\%& 1.8\%&83.1\% \\
    \hline
     \multicolumn{4}{c}{\textbf{EmoWOZ}}   \\
      \hline
      Fear & Diss & Apol & Abus \\ 
  
   0.5\%&6.1\%& 1.0\%& 0.2\%  \\

   & Exc & Sat & Neu \\
  &1.2\%& 21.0\%&70.1\% \\
    \hline
  \end{tabular} 
 \label{t1}
\end{table}

From Table  \ref{t0} it can be noted that EmoWOZ has almost double the amount of average turns per dialogue than DailyDialog. 

From Table \ref{t1} it can be observed that both datasets are imbalanced, not only for its dominant majority neutral class, but also for the relative imbalance between minority classes. Therefore, we have opted to use the macro-F1 score for evaluation in order to promote consistent performance across all classes.

%
%
%

\section{Results and Analysis}

\subsection{Iterating towards the ideal approach}

We have performed extensive experiments in order to obtain our ideal model architecture. From experimenting different approaches to pool the various layers of embeddings RoBERTa provides to choosing which classification module to employ withing a wide variety of deep learning architectures, we put forward our experiments in this subsection.

\vspace{2.5mm}
\subsubsection{Fine-tuning}

Fine-tuning, the modification of the pre-trained RoBERTa's weights along with the classification head during training with the target dataset, is a determinant procedure for the success of our approach. 

In our experiments we observed that if we did not fine-tune the language model and just trained the classification head, the model would always predict the majority neutral class.
This supports the notion that pre-trained-language models are useful for a wide variety of tasks but need to be fine-tuned for the specific task at hand.
\vspace{2.5mm}
\subsubsection{Pooling}

We have performed experiments with several pooling alternatives. From average pooling, max pooling, concatenation of the CLS token of more than 1 last layers to the concatenation of the CLS token with the result from average pooling. All these pooling alternatives resulted in lower performance than choosing the CLS token of the last layer. This might suggest a high representative power for the CLS token, which is proposed for classification, and discards the need for directly considering other tokens for this task.
\vspace{2.5mm}
\subsubsection{Classification module}

We have also performed alternative experiments with other classification modules than our simple classification head. These consisted in passing the pooled embeddings through Recurrent Neural Networks \cite{elman1991distributed}, uni \cite{hochreiter1997long} and bi-directional \cite{graves2005bidirectional} Long Short-Term Memory Networks and a Conditional Random Field  \cite{lafferty2001conditional} before feeding them to the classification head. Performance was lower in all alternative experiments when compared to our main approach of using a simple classification head. These results may indicate that our approach leveraging RoBERTa's representational power for context suffices and there is no apparent need for modelling the context with complex classification modules, after obtaining our context-dependent embedding utterance representations.

\subsection{Overall Performance}

For each of the datasets, we have performed experiments without introducing any context ($c=0$) to introducing 4 previous conversation turns ($c=4$), for which the overal performance operationalized by the macro-F1 metric is reported in Table \ref{t2}. Our results are an average of 5 runs.

\begin{table}[H]
 \centering
 \caption{Model performance in macro F1-score with the introduction of $c$ conversational turns}
  \begin{tabular}{cccc}
    \hline
       &\textbf{DailyDialog} & \textbf{EmoWOZ}\\
      &   macro-F1 & macro-F1\\
    \hline
    c=0&48.52&58.66\\
    c=1 &50.31&62.32\\
    c=2  & 50.44&64.98 \\
    c=3 &  \textbf{51.23}&\textbf{65.33}\\
    c=4& 50.46&63.28\\
    
    \hline
  \end{tabular} 
 \label{t2}
\end{table}

It can be observed that introducing previous conversational context turns leads to an increase in macro-F1 score. As hypothesised, providing no context is never the best option. This shows that the introduction of an adequate number of context turns directly as the language model input significantly improves model performance. In general performance increases with the introduction of each additional context turn up to the ideal number of turns and then it decreases. Overall, it can be concluded that the ideal number of introduced context turns for ERC in both datasets is 3.

\subsection{Performance on each emotion label}

For each dataset, we report the results on each individual emotion label and also present the confusion matrices for the best determined $c$ value. Our results are an average of 5 runs.

The individual emotion label F1-scores for the DailyDialog dataset are presented in Table \ref{t4}.

\begin{table*}[t]
 \centering
 \caption{Model performance on each individual emotion label on the \textbf{DailyDialog} dataset with the introduction of $c$ conversational turns}
  \begin{tabular}{cccccccc}
    \hline
     & \textbf{Ang} & \textbf{Disg} & \textbf{Fear} & \textbf{Hap} &\textbf{Sad} & \textbf{Sur} &\textbf{ Neu}   \\
    \hline
    c=0& 37.47 & 32.32 & 36.69 & 59.42 & 33.16 & 49.60 & 90.99    \\
    c=1 & 40.18 & 29.28 & 39.43 & 61.26 & 38.30 & \textbf{52.66} & 91.06 \\
    c=2& 43.26 & 33.91 & 36.52 & \textbf{61.98} & 33.63 & 52.23 & 91.12   \\
    c=3& \textbf{43.51} & 33.22 &\textbf{39.44} & 61.12 & \textbf{38.43} & 51.50 & \textbf{91.42} \\
    c=4&42.00 & \textbf{34.52} & 34.65 & 61.97 & 37.18 & 51.70 & 91.18 \\
    
    \hline
  \end{tabular} 
 \label{t4}
\end{table*}

\begin{table*}[t]
 \centering
 \caption{Model performance on each individual emotion label on the \textbf{EmoWOZ} dataset with the introduction of $c$ conversational turns}
  \begin{tabular}{cccccccc}
    \hline
     & \textbf{Fear} & \textbf{Diss} & \textbf{Apol} & \textbf{Abus} & \textbf{Exc} & \textbf{Sat} & \textbf{Neu}  \\
    \hline
    c=0 &35.72           &45.18 & \textbf{74.93}& 25.21& 46.96& 90.09& 92.53  \\
    c=1  &32.97          &57.97 &72.47 &42.97 &47.07 &89.75 & 93.01\\
    c=2 &\textbf{38.91} &66.24 &73.37 &44.79 &48.13 & 89.73&  93.74 \\
    c=3 &37.89 &68.02 & 72.49& \textbf{47.73}& 47.64&89.76 &93.81 \\
    c=4 &35.15 &\textbf{69.57} & 73.00&  30.09& \textbf{50.89}&\textbf{90.23}&\textbf{94.03} \\
    
    \hline
  \end{tabular} 
 \label{t6}
\end{table*}

It can be observed that for more than half of the labels, Anger, Fear, Sadness and Neutral, the ideal context to be provided is 3 turns which maximise their F1-scores, and also the macro-F1 score on Table \ref{t2}, and for the other labels the ideal context is 4 turns for Disgust, 2 turns for Happiness and 1 turn for Surprise. As expected, providing no context is never the best option. 

\begin{figure}[ht]
    \centering
   \includegraphics[width=0.98\linewidth]{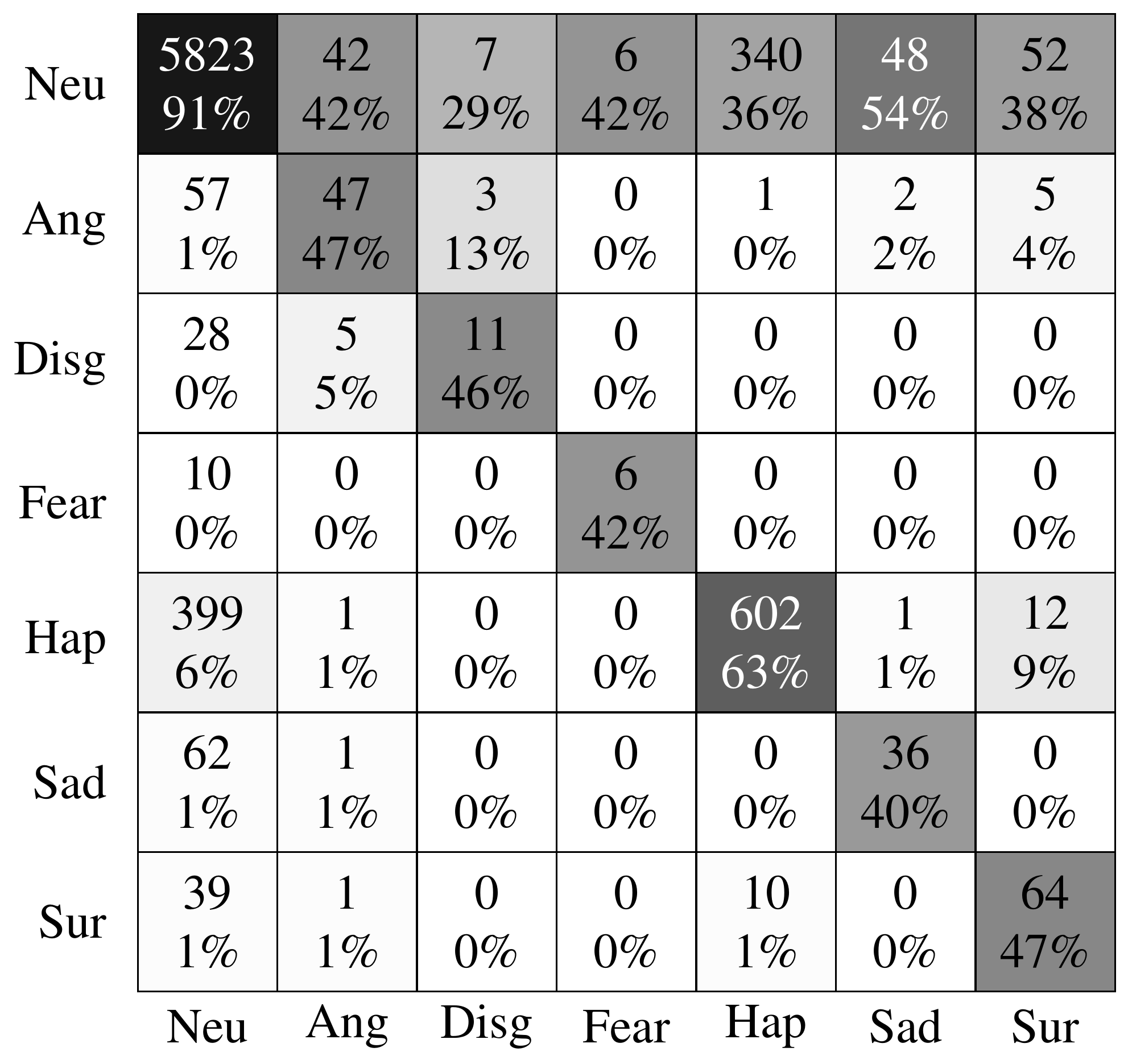}
\caption{Confusion Matrix for the  \textbf{DailyDialog} dataset with the introduction of $c$=3 conversational turns}
\label{c3}
\end{figure}

The confusion matrix for $c=3$ corresponding to the highest macro-F1 score is displayed on Figure \ref{c3}, in which the label nomenclature and order is the same as in table \ref{t4} but with neutral as the first label.

This matrix indicates that majority of the errors are due to classifying utterances as neutral instead of assigning a non-neutral emotion. The classifier also displays some confusion in discerning between Happiness and Surprised.

The individual emotion label F1-scores for the EmoWOZ dataset are presented in Table \ref{t6}.

It can be observed that for 4 of the labels, Dissatistfied, Excited, Satisfied and Neutral, the ideal context to be provided is 4 turns which maximise their F1-scores. Regarding the other labels the ideal context is 2 for Fear, 3 for Abusive, and surprisingly 0 turns for Apologetic, which might indicate that this emotion is very explicit in this dataset. 

\begin{figure}[ht]
    \centering
   \includegraphics[width=0.98\linewidth]{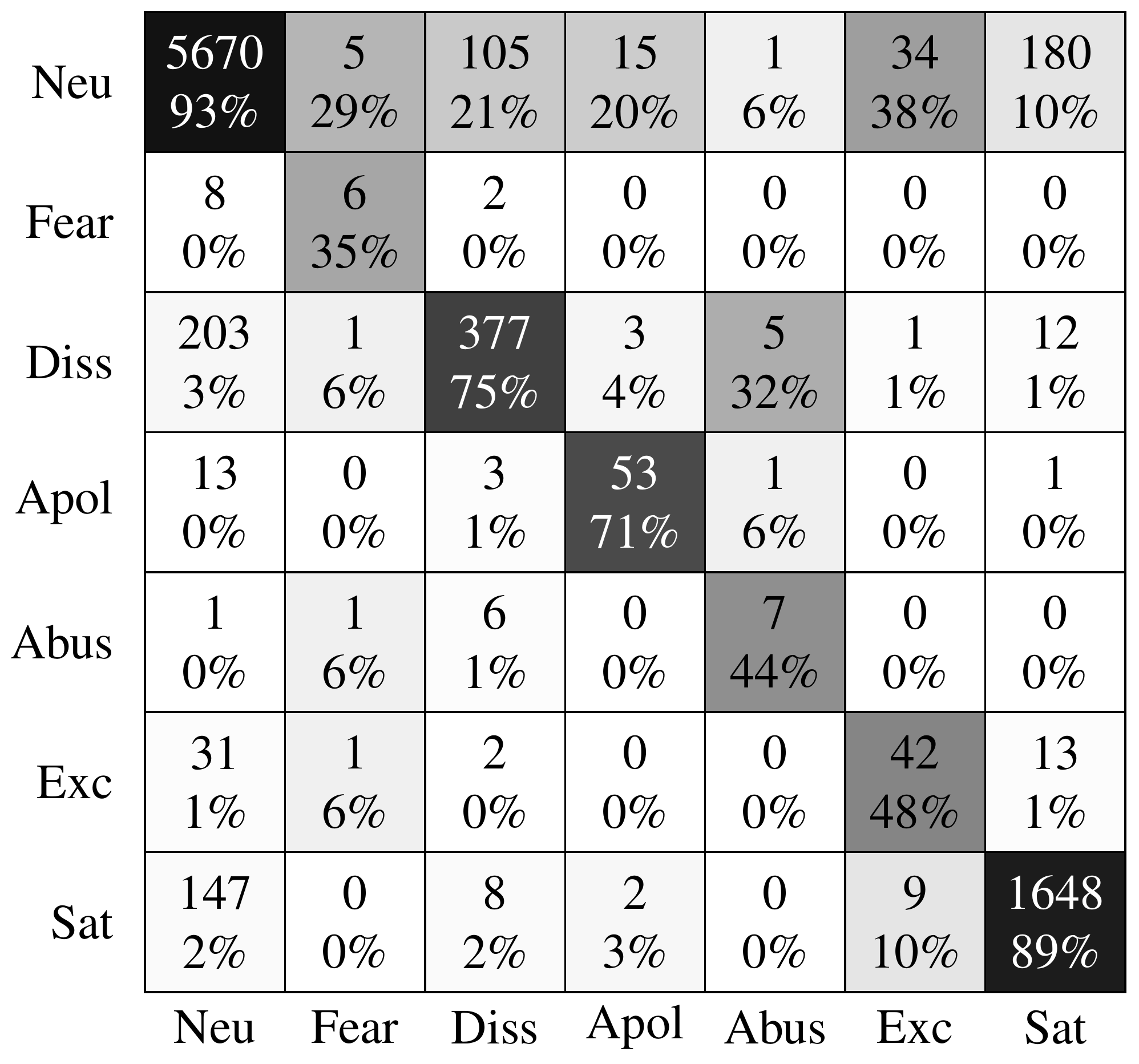}
\caption{Confusion Matrix for the  \textbf{EmoWOZ} dataset with the introduction of $c$=3 conversational turns}
\label{c2}
\end{figure}

The confusion matrix for $c=3$ corresponding to the highest macro-F1 score is displayed on Figure \ref{c2}, in which the label nomenclature and order is the same as in table \ref{t6} but with neutral as the first label. 

This matrix indicates that majority of the errors are due to classifying utterances as neutral instead of assigning a non-neutral emotion, as in happens with the DailyDialog dataset. 

\vspace{2.5mm}

It is worth noting that our results are an average of 5 runs and the final model is determined via performance on the validation set. Therefore, the fluctuation in individual label F1-scores does not hinder the representativity of our results and these fluctuations may occur between results from the other reported state-of-the-art models.

\subsection{Comparison with state-of-the-art}
\begin{table*}[t]
 \centering
 \caption{Comparison with state-of-the-art works}
  \begin{tabular}{cccc}
    \hline
       &\textbf{DailyDialog} & \textbf{EmoWOZ}\\
       &macro-F1 & macro-F1 \\
    \hline
    RoBERTa \cite{ghosal-etal-2020-cosmic} / BERT \cite{feng2022emowoz} &48.20&55.80\\
    RoBERTa \cite{ghosal-etal-2020-cosmic} / BERT \cite{feng2022emowoz} + DlgRNN  &49.65&57.10\\
    ContextBERT \cite{feng2022emowoz} &-&59.70\\
    COSMIC \cite{ghosal-etal-2020-cosmic} / \cite{feng2022emowoz}&51.05&61.12\\
    Psychological \cite{li-etal-2021-past-present}&\textbf{51.95}&-\\
    \hline
    CD-ERC (Ours)&51.23&\textbf{65.33} \\

    \hline
  \end{tabular} 
 \label{t5}
\end{table*} 

We further compare our approach to other state-of-the-art approaches that also resort to the RoBERTa or BERT pre-trained-language models. This allows for a fair comparison between approaches given that using this language model brings great performance increases when compared to using other means of utterance feature extraction. Regarding DailyDialog results, we compare our approach to COSMIC \cite{ghosal-etal-2020-cosmic}, RoBERTa and RoBERTa DialogueRNN, implemented by the authors of COSMIC, and the Psychological model \cite{li-etal-2021-past-present}, all models described in Section \ref{rl}. Concerning the performance on the EmoWOZ dataset, we compare out approach to COSMIC, BERT and BERT DialogueRNN, tested by the authors of EmoWOZ \cite{feng2022emowoz}, since for this dataset the authors obtained a more suitable uterrance representation using BERT instead of RoBERTa. Results are displayed on table \ref{t5} and are an average of 5 runs.

Regarding performance on the DailyDailog dataset, our approach outperforms not only the simple RoBERTa/BERT, but also RoBERTa/BERT in a more elaborate gated neural network model such as DialogueRNN and COSMIC. The Psychological model has a slightly higher performance than ours. It may be due to the fact that it leverages a large commonsense knowledge base and an elaborate classifier architecture, while we opted for a minimalistic classification module.
Concerning performance on the EmoWOZ dataset, our approach outperforms all baselines by a wide margin, setting a new state of the art for task-oriented emotion datasets.

\subsection{Case Studies}

On Table \ref{tcsdd} we can compare the performance of our contextual classifier when considering the ideal 3 context turns on both datasets versus not considering any context at all. 

\begin{table*}[htpb]
 \centering
 \caption{Case studies comparing the performance of our contextual classifier ($c=ideal=3$) with the no-context classifier ($c=0$)}
  \begin{tabular}{ccccc}
    \hline
       Turn & Gold & $c=0$ & $c=3$\\
      \hline
      A: Can I help you ? & Neu& Neu&Neu \\
      B: I would actually like to view the apartment for rent today .& Neu& Neu &Neu \\
      A: \textbf{I ’ m sorry , but you won ’ t be able to view it today .}& \textbf{Sad}& \textbf{Neu} & \textbf{Sad}\\
       \hline 
      
      A: Maybe you should look around for an outlet .& Neu&Neu&Neu\\
      B: That is a wonderful idea . &Hap&Hap&Hap\\
    A: Outlets have more reasonable prices . &Neu&Neu&Neu\\
    B: Thank you for your help .& Hap&Hap&Hap\\
    A: \textbf{No problem . Good luck} &\textbf{Hap}&\textbf{Neu}&\textbf{Hap}\\
    \hline
      A: On what day would you like to travel? &- & - & -\\
      B: \textbf{Saturday, please. I'm thinking just a short vacation over the weekend.}&  \textbf{Neu} &  \textbf{Exc} &  \textbf{Neu}\\

    \hline
  \end{tabular} 
 \label{tcsdd}
\end{table*}

In the first example, from the DailyDialog dataset, A offers B assistance, so B asks A to view the apartment, to which A sadly apologizes informing B that B will not be able to view it. The classifier that does not consider context classifies this last apology as neutral. However, given the context of the conversation, A should not be neutral since A is unable to assist B which was A's initial purpose. The contextual classifier is able to consider this, thus correctly classifying A's utterance with the emotion Sadness.  

In the second example, also from the DailyDialog dataset, A gives B a good idea to which B happily reacts and thanks A. A happily reacts to B's acknowledgments, especially since B mentioned A's was a "wonderful idea". The classifier that does not consider context classifies A's final reaction to B as neutral, since A's utterance is a merely "No problem. Good luck", not being able to recognize A's positive reaction to B's acknowledgements. The contextual classifier, however, having this utterances into account, correctly classifies A's final reaction with the emotion Happiness.

In the last example, from the EmoWOZ dataset, B is merely answering A's question of what day B would like to travel. The classifier that does not consider context takes into account the words "please" and "vacation" which bias the classification towards the emotion Excited. The contextual classifier might grasp that "please" is used as a polite expression and "vacation" is just the object of the phrase, thus correctly classifying the utterance as neutral.

\section{Conclusions and Future Work}

In this work we have leveraged context-dependent embedding utterrance representations for Emotion Recognition in Conversations. Our approach of producing context-dependent representations of each utterance contrasted with the usual approach of producing context independent representations of each utterance and subsequently performing contextual modeling of these. It consisted in feeding a variable number of previous conversational turns appended to the utterance to be classified as input to the state-of-the-art pre-trained-language model RoBERTa, to which we appended a simple classification module. We further investigated how the number of introduced conversational turns influenced our model performance.
We concluded that the introduction of an adequate number of context turns directly as the language model input significantly improves model performance.

Furthermore, we attained state-of-the-art results on the widely used DailyDialog dataset and established a new state-of-the-art by a wide margin on the EmoWOZ dataset, which are usually yielded by more elaborate classifiers resorting to larger state-of-the-art pre-trained-language models and more complex classification modules. 

For future work, from adequately capturing the conversation context, the focus of our approach, to capturing several other factors that influence the emotions in the conversation, such as self and inter-speaker emotional influence and the emotion of preceeding utterances, various architectures comprising not only state-of-the art language models for embeddings but also combining our context-dependent embedding utterance representation with more elaborate classification modules can be used.

Finally, we put forward important ethical aspects pertaining to Emotion Recognition in Conversations. These are, for example and not limited to, whether an ERC module should be developed or used for a certain purpose, which data to collect, the subjects behind the data, diversity, inclusiveness, privacy, control and possible biases and misuses of the application \cite{mohammad2022ethics}. Research taking into account these aspects will benefit the community with better ERC modules for current and novel applications.

\section{Acknowledgements}

This work was supported by Fundação para a Ciência e a Tecnologia (FCT), through Portuguese national funds Ref. UIDB/50021/2020, Agência Nacional de Inovação (ANI), through the project CMU-PT MAIA Ref. 045909, RRP and Next Generation EU project Center for Responsible AI Ref. C645008882-00000055, and the COST Action Multi3Generation Ref. CA18231.

\bibliography{anthology,custom}
\bibliographystyle{acl_natbib}

\end{document}